\documentclass{article}

    \PassOptionsToPackage{numbers, compress}{natbib}

 \usepackage[main, final]{neurips_2026}

\usepackage[utf8]{inputenc} 
\usepackage[T1]{fontenc}    
\usepackage{hyperref}       
\usepackage{url}            
\usepackage{booktabs}       
\usepackage{amsfonts}       
\usepackage{nicefrac}       
\usepackage{microtype}      
\usepackage{xcolor}         
\usepackage{amsmath}
\usepackage{color-edits}
\usepackage{float}
\usepackage{adjustbox}
\usepackage{multirow}
\usepackage{colortbl}
\usepackage{enumitem}
\usepackage{fontawesome}
\usepackage{algorithm}
\usepackage{algpseudocode}
\addauthor[Mahdi]{m}{blue}

\title{Model-Adaptive Tool Necessity Reveals the Knowing-Doing Gap in LLM Tool Use}

\author{%
  Yize Cheng\thanks{Equal contribution} \quad
  Chenrui Fan\footnotemark[1] \quad
  Mahdi JafariRaviz\footnotemark[1] \quad
  Keivan Rezaei \quad
  Soheil Feizi \\
  University of Maryland, College Park \\
  \texttt{\{yzcheng, cfan42, krezaei, mahdij, sfeizi\}@umd.edu}\\
  \\
  \faGithub~\textbf{Project}: \url{https://github.com/chengez/Tool-Cognition-Action}
}

\begin{document}

\maketitle

\begin{abstract}

Large language models (LLMs) increasingly act as autonomous agents that must decide when to answer directly vs. when to invoke external tools. Prior work studying adaptive tool use has largely treated tool necessity as a model-agnostic property, annotated by human or LLM judge, and mostly cover cases where the answer is obvious (e.g., fetching
the weather vs.\ paraphrasing text). 
However, tool necessity in the wild is more nuanced due to the divergence of capability boundaries across models: a problem solvable by a strong model on its own may still require tools for a weaker one.
In this work, we introduce a model-adaptive definition of tool-necessity, grounded in each
model's empirical performance. Following this definition, we compare the necessity against observed tool-call behavior across four models on arithmetic and factual QA dataset, and find substantial mismatches of 26.5–54.0\% and 30.8–41.8\%, respectively.
To diagnose the failure, we decompose tool use into two stages: an internal cognition stage that reflects whether a model believes a tool is necessary, and an execution stage that determines whether the model actually makes a tool-call action. 
By probing the LLM hidden states, we find that both signals are often linearly decodable, yet their probe directions become nearly orthogonal in the late-layer, last-token regime that drives the next-token action. 
By tracing the trajectory of samples in the two-stage process, we further discover that the majority of mismatch is concentrated in the cognition-to-action transition, not in cognition itself. These results reveal a \emph{knowing--doing gap} in LLM tool-use: improving tool-use reliability requires not only better recognition of when tools are needed, but also better translation of that recognition into action.
\end{abstract}
\section{Introduction}
\label{sec:intro}
Large language models (LLMs) are increasingly deployed as autonomous agents that interact with external tools such as search engines, calculators, and APIs~\citep{parisi2022talm, schick2023toolformer, song2023restgpt, mialon2023augmented}. A central challenge in building reliable autonomous LLM agents is achieving adaptive tool using: the LLM needs to determine \textit{when} it should rely on such tools versus answering directly~\citep{huang2024metatoolbenchmarklargelanguage, qian2025smartselfawareagenttool,wang2026positionagentinvokeexternal}. Prior work studying adaptive tool use~\citep{huang2024metatoolbenchmarklargelanguage,qian2025smartselfawareagenttool,li2025adaptive} has largely treated tool necessity as a static, model-agnostic property, typically relying on human annotators or strong LLM judges to determine whether a query requires a tool, focusing primarily on polarized cases where the answer is obvious, such as fetching real-time weather data versus paraphrasing a static paragraph. 
However, tool necessity in the wild is fundamentally more nuanced due to the natural divergence of capability boundaries across different models. A problem that is easily solvable by a state-of-the-art model relying solely on its internal weights may completely exceed the capabilities of a smaller or less capable model, thereby making tool use strictly necessary for the latter but redundant for the former.

In this work, we argue that tool necessity must be intrinsically tied to the specific capabilities of the model in question. We introduce a model-adaptive definition of tool necessity, grounded not in static annotations, but in each individual model's empirical performance. By evaluating necessity relative to a model's intrinsic capabilities, we establish a more accurate characterization for when a specific LLM should seek external help. Following this definition, we compare the actual necessity against the observed tool-call behavior across four distinct models on arithmetic and factual question-answering (QA) datasets. Our findings reveal substantial mismatches: models exhibit a 26.5–54.0\% necessity-action mismatch in arithmetic tasks and a 30.8–41.8\% necessity-action mismatch in factual QA, frequently calling tools when capable of answering directly, or attempting to answer directly when lacking the requisite internal knowledge.

\begin{figure}[t]
    \centering
    \includegraphics[width=\linewidth]{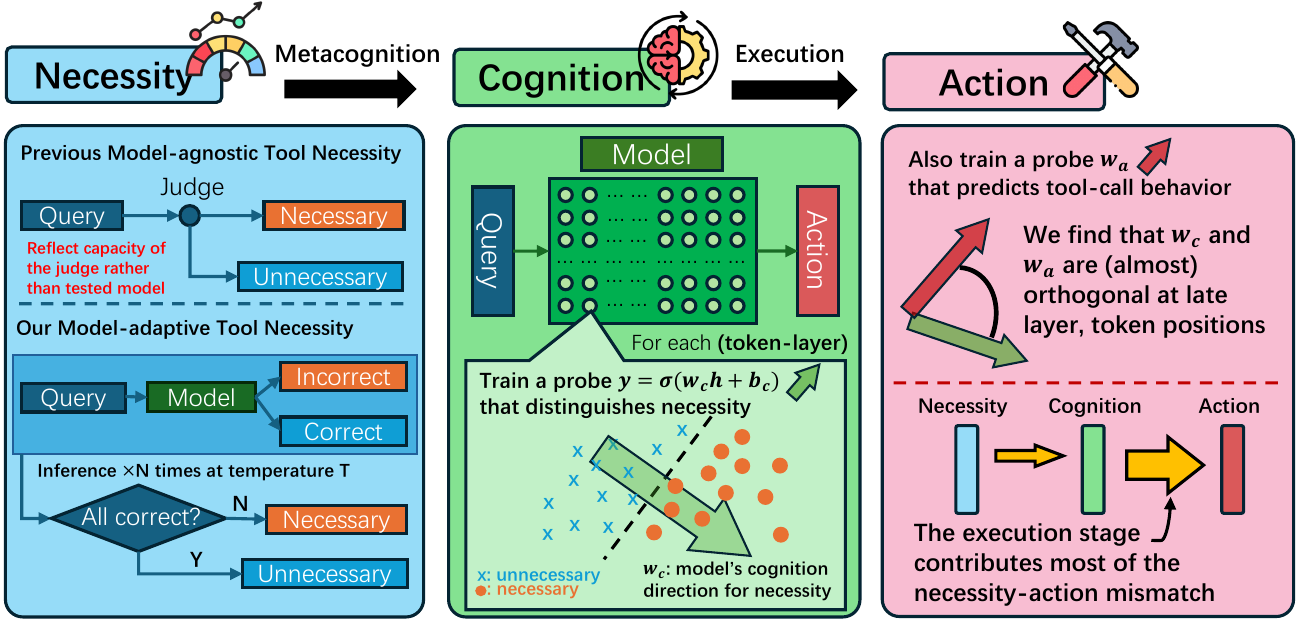}
    \vspace{-0.5cm}
    \caption{\textbf{Overview of the two stage cognition-execution modeling of LLM tool-use.} \textbf{(Left) Necessity}: We introduce a model-adaptive definition of tool necessity based on a model's empirical ability to consistently answer a query correctly on its own, contrasting with prior model-agnostic approaches. \textbf{(Middle) Cognition}: By probing the model's internal hidden states $h$, we identify a linear cognition direction $w_c$ that successfully distinguishes when a tool is necessary. \textbf{(Right) Action}: We also train a probe $w_a$ to predict the actual tool-call execution. We find that $w_c$ and $w_a$ become nearly orthogonal in late layers, and that the majority of the necessity-action mismatch stems from the execution stage (translating awareness into action) rather than the internal cognition stage.}
    \vspace{-7pt}
    \label{fig:teaser}
\end{figure}

To diagnose the underlying mechanisms of this failure, we propose a two-stage decomposition of the tool-use process: an internal cognition stage, which reflects whether the model's internal representations encode the belief that a tool is necessary, and an execution stage, which determines whether the model actually outputs the tool-triggering tokens. Building on prior advancements in mechanistic interpretability and representation engineering~\citep{zou2025representationengineeringtopdownapproach} and following recent literature on adaptive tool-using~\citep{li2025adaptive,wang2026asatrainingfreerepresentationengineering}, we probe the LLM hidden states and find that both the cognition of necessity and the execution intent are often linearly decodable. Yet, intriguingly, their respective probe directions become nearly orthogonal in the late-layer, last-token regime.

By tracing the trajectory of samples through this two-stage process, we uncover a \textit{knowing-doing gap} in LLM tool use: the majority of the observed necessity-action mismatch cases originates from the transition from cognition to action, rather than in the cognition stage. Models frequently generate internal representations indicating the awareness of their own limitations, but fail to translate this into the syntactic execution of a tool call. Our main contributions can be summarized as follows:
\begin{itemize}[leftmargin=*]
    \item We introduce a model-adaptive definition of tool necessity grounded in empirical performance, challenging the traditional reliance on static, model-agnostic annotations.
    \item We evaluate four distinct LLMs across arithmetic and factual QA datasets, revealing substantial behavioral mismatches (up to 54.0\%) between actual tool necessity and observed tool-call actions.
    \item By dividing tool use into an internal cognition stage and an execution stage, we use representation probing to demonstrate that while both intent and necessity are linearly decodable, their probe directions become near orthogonal in the late-layer, last-token regime.
    \item Through trajectory tracing, we discover that tool-use failures predominantly occur during the transition from cognition to action, highlighting a knowing-doing gap in LLM tool-use.
\end{itemize}

\section{Related work}

\paragraph{Tool calling in LLM agents.}
To extend LLM capabilities beyond parametric knowledge, researchers have introduced function/tool calling~\citep{parisi2022talm, schick2023toolformer, song2023restgpt, mialon2023augmented}, enabling interaction with external resources and expanding task coverage. Standardized protocols like MCP~\citep{anthropic2024mcp} and A2A~\citep{google2025a2a} further streamline communication and access within tool ecosystems.
In parallel, various works has examined tool-use accuracy~\citep{li2023api, patil2025berkeley}, hallucinated calls~\citep{zhang2024toolbehonest, ross2025when2call}, and robustness to tool descriptions~\citep{shi2025prompt, faghih2025gaming}. However, while these efforts aim at teaching and evaluating \textit{how} to use tools, an important and often understudied challenge in building reliable LLM agents is determining \textit{when} to use tools. Existing works that do study this challenge~\citep{huang2024metatoolbenchmarklargelanguage, qian2025smartselfawareagenttool,li2025adaptive} treat tool necessity as a static property of the query, labeling instances as either tool-necessary or tool-unnecessary using human annotators or some proprietary LLM. This ignores the inherent difference in capability boundaries between different models. While \citet{wang2026positionagentinvokeexternal} has also advocated for model-dependent tool necessity, to the best of our knowledge, we are the first to have a pipeline that empirically grounds tool necessity in the actual capabilities of a given model.

\paragraph{Meta-cognition of LLMs and the ``knowing-doing gap''.}
The ability of LLMs to accurately assess their own capability boundaries—often referred to as meta-cognition or self-assessment—has been a topic of long-standing interest~\citep{kadavath2022languagemodelsmostlyknow, yin-etal-2023-large}. To measure this self-awareness, early work primarily relies on measuring explicit self-assessment by teaching models to express their knowledge boundaries~\citep{chen2024teachinglargelanguagemodels,zhang2024rtuninginstructinglargelanguage} or to directly verbalize confidence~\citep{lin2022teachingmodelsexpressuncertainty}.
However, recent work has shown that the ability for models to verbalize its internal activations is limited~\citep{lindsey2025biology,jian2025languagemodelscapablemetacognitive}. Moreover, the task of self-assessment and actual problem solving are fundamentally different tasks. When explicitly prompted about its capability boundary, the model would focus on self-assessment. But when faced with actual problem solving, the prompt is usually tasks-oriented, and hence the self-assessing process becomes implicit and subconscious. This akin to the distinction between system I and system II thinking~\citep{li202512surveyreasoning}. Therefore, in this work, we follow some recent work that use internal state probing to measure models' cognition of tool-necessity~\citep{li2025adaptive,wang2026asatrainingfreerepresentationengineering}, and also empirically show in Appendix~\ref{append:verbal_tool_necessity} how model tool-call actions change when explicitly prompted for self-assessment. 

Meanwhile, papers in other domain of LLMs that leverage hidden states to study model internal cognition have found that the model's action can diverge from its internal belief. For example, \citet{zhao2025llmsencodeharmfulnessrefusal} find that LLMs may fail to refuse harmful queries despite internally recognizing their harmfulness, and \citet{zhang2026stopfailoperationalcapability} show that models can internally recognize their inability to solve certain math problems yet still expend tokens on unproductive reasoning. In this work, we show that this ``knowing-doing gap'' similarly exists in tool-calling, and it can constitute even a larger proportion of end-to-end errors.  




\section{Defining model-adaptive tool necessity and two-stage modeling of tool-call}
\label{sec:formulation}

To study tool-use behavior in LLMs, we introduce a simple decomposition that separates recognizing the need for a tool from acting on that recognition. This distinction will serve as the foundation for the evaluation, diagnosis, and analysis throughout the rest of this paper.

\paragraph{Defining model-adaptive tool necessity.}
Existing work typically assumes a fixed notion of tool necessity, assigning each query a static label independent of the model being evaluated. However,  we argue that since different models have different capability boundaries, the tool necessity label should be adaptive according to the model.
To characterize a model's capability boundary, given a model $f$ and query $x$, we perform $N$ independent inference runs without access to external tools at temperature $T$. 
If the model $f$ can consistently solve the problem $x$ correctly across $N$ runs, we assume that this $x$ falls within the $f$'s capability boundary and therefore the tool necessity, $n_f(x)$, is $0$. Otherwise, the model cannot reliably solve this query, and hence $n_f(x)$ is $1$. The parameters $N$ and $T$ control the strictness of this criterion. Specifically, larger values of $N$ and $T$ yield a more conservative and robust estimate of whether a query truly falls within the model’s capability boundary as they demand the model to output the correct answer more consistently.

This formulation captures a key aspect of real-world deployment: reliability under uncertainty. In practical settings, a model that only occasionally produces the correct answer without tools may still benefit from external assistance to ensure consistent performance. By grounding tool necessity in empirical behavior rather than static annotation, our approach provides a more faithful characterization of when tool use is genuinely required for a given model.

\paragraph{The cognition-execution modeling of tool-call.}
We conceptualize tool use as a two-stage process:
\begin{equation}
    x \rightarrow z_f(x) \rightarrow a_f(z_f(x)),
\label{eq:two-stage-modeling}
\end{equation}
where $z_f(x)$ represents the model's internal cognition of whether a tool is needed, and $a_f(z_f(x))$ denotes whether the model actually invokes a tool, based on its cognition.
This two-stage decomposition mirrors the cognition process of human and what we desire for the model.
It distinguishes between \textit{meta-cognition}—the model's internal belief about its capability boundary, and \textit{execution ability}—how model acts based on its cognition.

\paragraph{End-to-end error diagnosis.}
Under our model-dependent definition of tool necessity $n_f(x)$ and the two stage modeling as in Equation~\ref{eq:two-stage-modeling}, we can decompose the end-to-end necessity-action mismatch, $D(n_f(x), a_f(z_f(x)))$, into the mismatch between actual necessity and cognition $D(n_f(x),z_f(x))$, and the mismatch between model's cognition and actual decision $D(z_f(x), a_f(z_f(x))$, where $D(m,n)$ denotes the discrepancy between $m$ and $n$.


\section{Dataset curation}

We cover two representative domains: math arithmetic and factual question answering, using two widely used model families: Qwen3-8B and Qwen3-4B \citep{yang2025qwen3technicalreport}, as well as Llama-3.1-8B-Instruct and Llama-3.2-3B-Instruct \citep{grattafiori2024llama}. These domains provide natural testbeds in which some queries can be reliably solved by the model alone, while others may require external assistance (i.e. a calculator for arithmetic tasks and a search API for factual queries).
For math arithmetic dataset, we mix problem types that vary in both surface form and actual difficulty. It includes simple one- and two-step addition and subtraction problems, along with harder examples involving multi-digit multiplication, modulo, parentheses, operator precedence, and longer addition/subtraction chains, resulting in a total of 4,000 instances.
This gives us problems with a range of difficulty levels from very simple questions to extremely difficult ones, enabling us to measure the capability boundary of the model.
More details about the curation of our arithmetic dataset can be found in Appendix~\ref{app:arithmetic}.
For factual question answering, we adopt TruthfulQA~\citep{lin2022truthfulqameasuringmodelsmimic}, a widely used dataset with 817 instances designed to evaluate the factual reliability of language models. 

\begin{figure}[h]
    \centering
    \includegraphics[width=1\linewidth]{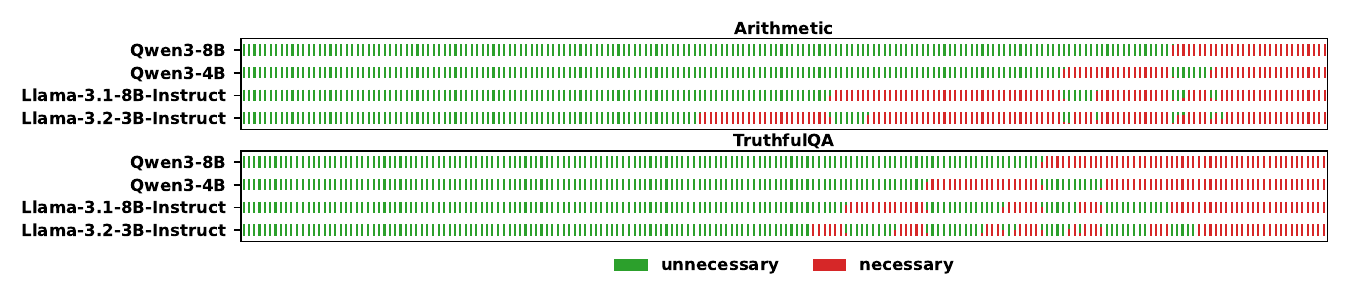}
    \vspace{-0.6cm}
    \caption{\textbf{Model-adaptive tool-call necessity.} Each vertical bar represents 0.5\% of samples. Green indicates samples answered correctly in all $N=10$ no-tool runs; red indicates at least one failure. Within each dataset, samples share the same order across rows, obtained by recursively sorting within each previous model's correctness partition.}
    \vspace{-0.4cm}
    \label{fig:bound}·
\end{figure}

\subsection{Grounding tool necessity to model-specific capability boundaries}
\label{sec:necessary_and_unnecessary}

We follow our definition in Section~\ref{sec:formulation} and run $N = 10$ independent inferences at temperature $T = 0.7$ without access to external tools. Sensitivity analysis under alternative $N$, $T$ values are presented in Appendix~\ref{append:alternative_NT}. For a specific model, we count samples where the model fails at least once as \emph{tool-necessary}, and samples where the model consistently gives correct answers across all $N=10$ runs as \emph{tool-unnecessary}. Figure~\ref{fig:bound} shows that different models have substantially different capability boundaries, which would be obscured by the model-agnostic definition of tool necessity. Specifically, the clean boundary in the first row is induced by our sorting procedure, while the red-green disagreements across rows show that the same sample groups can fall on different sides of different models' capability boundaries. This pattern appears in both arithmetic and factual question answering, suggesting that tool necessity depends not only on task type or dataset membership, but also on the particular model being deployed. This motivates using $n_f(x)$ rather than a single global necessity label when evaluating tool-use judgment and downstream call behavior.

\subsection{Collecting tool-call behaviors on \emph{tool-necessary} and \emph{tool-unnecessary} instances}
\label{sec:collect_tool_Call_behav}
We run inference on the LLMs using both the tool-necessary and tool-unnecessary instances obtained in Section~\ref{sec:necessary_and_unnecessary}. In this setting, models are provided access to external tools: a calculator for arithmetic question answering and a search API for factual queries. To facilitate the diagnostic interpretation efforts in Section~\ref{sec:from_meta_to_exe_what_went_wrong}, greedy decoding is used when collecting tool-call actions. To better reflect real-world deployment, we follow existing practice~\citep{patil2025berkeley,cheng2026llmagentstemporallyblind} and implement model-specific handlers that expose these tools in the syntax expected by each model. 
We then further divide \emph{tool-necessary} and \emph{tool-unnecessary} samples based on the model's actual tool-call behavior, and obtain 4 sets of data: \emph{Necessary-Called} (N-C), \emph{Necessary-NotCalled} (N-NC), \emph{Unnecessary-Called} (UN-C), and \emph{Unnecessary-NotCalled} (Un-NC). The first and last are aligned with the optimal behavior under our model-dependent definition of necessity, while the middle two correspond to the end-to-end necessity--action mismatch $D(n_f(x), a_f(z_f(x)))$ defined in Section~\ref{sec:formulation}.
\begin{table}[ht]
    \centering
    \caption{Breakdown of tool-call behavior across the four categories defined by the model-dependent necessity $n_f(x)$ and the observed action. Aligned cells (\textcolor{green!50!black}{N-C}: Necessary-Called, \textcolor{green!50!black}{UN-NC}: Unnecessary-NotCalled) are shaded green; misaligned cells (\textcolor{red!70!black}{N-NC}, \textcolor{red!70!black}{UN-C}) are shaded red and together form the end-to-end mismatch, summarized in the gray \textbf{Mis.}\ column.}
    \label{tab:call_categories}
    \resizebox{\linewidth}{!}{%
    \begin{tabular}{l>{\columncolor{green!15}}c>{\columncolor{red!15}}c>{\columncolor{red!15}}c>{\columncolor{green!15}}c>{\columncolor{black!12}}c>{\columncolor{green!15}}c>{\columncolor{red!15}}c>{\columncolor{red!15}}c>{\columncolor{green!15}}c>{\columncolor{black!12}}c}
    \toprule
    \multirow{2}{*}{Model} & \multicolumn{5}{c}{Arithmetic} & \multicolumn{5}{c}{TruthfulQA} \\
    \cmidrule(lr){2-6} \cmidrule(lr){7-11}
     & N-C & N-NC & UN-C & UN-NC & \textbf{Mis.} & N-C & N-NC & UN-C & UN-NC & \textbf{Mis.} \\
    \midrule
    Qwen3-8B & 438 (11.0\%) & 140 (3.5\%) & 1526 (38.2\%) & 1896 (47.4\%) & \textbf{41.7\%} & 69 (8.4\%) & 146 (17.9\%) & 108 (13.2\%) & 494 (60.5\%) & \textbf{31.1\%} \\
    Qwen3-4B & 253 (6.3\%) & 581 (14.5\%) & 481 (12.0\%) & 2685 (67.1\%) & \textbf{26.5\%} & 103 (12.6\%) & 153 (18.7\%) & 189 (23.1\%) & 372 (45.5\%) & \textbf{41.8\%} \\
    Llama-3.1-8B-Instruct & 419 (10.5\%) & 1204 (30.1\%) & 335 (8.4\%) & 2042 (51.1\%) & \textbf{38.5\%} & 98 (12.0\%) & 130 (15.9\%) & 122 (14.9\%) & 467 (57.2\%) & \textbf{30.8\%} \\
    Llama-3.2-3B-Instruct & 526 (13.2\%) & 1559 (39.0\%) & 600 (15.0\%) & 1315 (32.9\%) & \textbf{54.0\%} & 58 (7.1\%) & 164 (20.1\%) & 104 (12.7\%) & 491 (60.1\%) & \textbf{32.8\%} \\
    \bottomrule
    \end{tabular}%
    }
\end{table}

\paragraph{End-to-end mismatch is substantial.}
Table~\ref{tab:call_categories} reports the distribution of the four categories across four models and two domains. The aggregated mismatch rate (gray \textbf{Mis.}\ column) ranges from \textbf{26.5\%} to \textbf{54.0\%} on arithmetic and from \textbf{30.8\%} to \textbf{41.8\%} on TruthfulQA, indicating that with a model-specific notion of tool necessity, between roughly one quarter and one half of all queries result in a tool-use action that is inconsistent with the model's actual capability. This mismatch rate between actual tool necessity and model tool-use action further highlights the importance of determining \textit{when} to use tools, an issue that is often overlooked in prior work that only emphasizes \textit{how} to use them.


\paragraph{The dominant failure mode is highly model- and domain-dependent.}
Beyond the overall mismatch rates, the specific types of errors vary significantly across both models and domains. On arithmetic, Qwen3-8B suffers from tool-overuse (UN-C at \textbf{38.2\%} vs. N-NC at \textbf{3.5\%}). In contrast, Qwen3-4B and both Llama models exhibit clear tool underuse, with N-NC rates of \textbf{14.5\%} (Qwen3-4B), \textbf{30.1\%} (Llama-3.1-8B-Instruct), and \textbf{39.0\%} (Llama-3.2-3B-Instruct), exceeding their respective UN-C rates. Interestingly, these tendencies are not consistent even within a single model. On TruthfulQA, Qwen3-8B reverses its trend entirely, showing tool underuse (N-NC at \textbf{17.9\%} vs. UN-C at \textbf{13.2\%}), while Qwen3-4B now shows tool-overuse (UN-C at \textbf{23.1\%} vs. N-NC at \textbf{18.7\%}). Because these models shift between being overly eager and overly conservative in tool-calling depending on the context, it is clear that no single, uniform bias can fully explain these mismatch errors. Therefore, in the next section, we leverage our two-stage modeling of LLM tool-use defined in Section~\ref{sec:formulation} for more fine-grained diagnosis.  


\section{From meta-cognition to execution ability: What went wrong?}
\label{sec:from_meta_to_exe_what_went_wrong}
Having measured the model-dependent tool necessities for each model (i.e., their capability boundaries) and collected their actual tool-call behaviors, we now examine where the breakdown between actual necessity and final action occurs, following the two-stage decomposition in Section~\ref{sec:formulation}. We first show that each stage—the internal cognition of necessity, and the executed action—is individually linearly separable from the model's hidden states (Section~\ref{sec:probe_cognition} and Section~\ref{sec:probe_action}), and then characterize the geometric relationship between the two (Section~\ref{sec:cognition_execution_gap}). Finally, we find that the majority of the error originates in the execution stage through per sample tracing (Section~\ref{sec:2_stage_error_diagnosis}).

\subsection{Probing for model's cognition}
\label{sec:probe_cognition}

Linear probing is a standard method for studying how concepts are represented in a model's hidden-state space. Recent works~\citep{li2025adaptive, wang2026asatrainingfreerepresentationengineering} have used it as a proxy for models' internal belief of tool necessity and reported that, despite substantial end-to-end mismatch, the hidden states of \emph{tool-necessary} and \emph{tool-unnecessary} samples are \textit{almost linearly separable}. Because that conclusion was drawn under a static, query-only definition of tool necessity, it is unclear whether it survives the model-dependent definition introduced in Section~\ref{sec:formulation}, where the necessity label $n_f(x)$ varies across models with different capability boundaries.

\begin{figure}[h]
    \centering
    \includegraphics[width=1\linewidth]{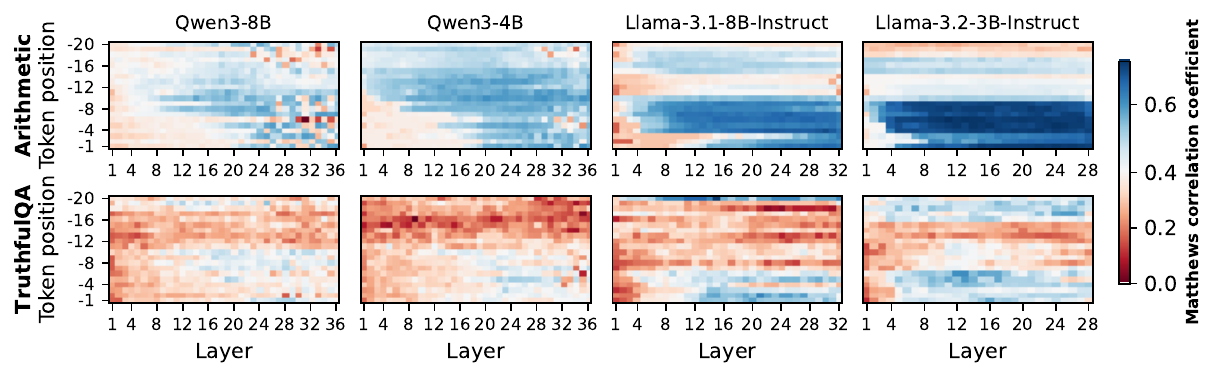}
    \vspace{-0.3cm}
    \caption{\textbf{Necessity probe performance across token-layer positions.} Each cell reports the held-out MCC of a linear probe trained to predict the model-adaptive necessity from the hidden state at a given layer and token position. Darker blue indicates stronger linear separability. The linear separability is strongly task-dependent, and the heatmap structure appears similar within model families.}
    \label{fig:probe-necessity}
\end{figure}

Concretely, we train a linear classifier with weight $\mathbf{w}_c$ and bias $b_c$ on the model's hidden states, using a learning rate of $0.01$ with the Adam~\citep{kingma2017adammethodstochasticoptimization} optimizer, minimizing the following objective:
\begin{equation}
     \mathcal{L}=-\frac{1}{K} \sum_{k=1}^{K} \left[ n_f(x_k)  \log \sigma(\mathbf{w}_c^\top h_t^{(l)}(x_k)+b_c) + (1 - n_f(x_k)) \log (1 - \sigma(\mathbf{w}_c^\top h_t^{(l)}(x_k)+b_c)) \right],
\label{eq:probe_objective}
\end{equation}
where $x_k$ is a sample in the dataset and $h_t^{(l)}(x_k)$ is the hidden state at token position $t$ and layer $l$. $\mathbf{w}_c$ also serves as the normal vector of the separating hyperplane, indicating the direction from ``unnecessary'' to ``necessary'' in the model's representation space. We sweep $(t, l)$ over all layers and over the last $20$ query tokens; negative indices denote token positions relative to the start of generation, e.g., $t=-1$ is the final query token.
As the class distribution is imbalanced (Table~\ref{tab:call_categories}), we report the probe performance using the Matthews Correlation Coefficient (MCC)~\citep{MATTHEWS1975442} on the held-out test set (30\% of data), which is a more robust metric than accuracy or F1 under skewed labels:
\begin{equation}
    \text{MCC} = \frac{TP \cdot TN - FP \cdot FN}
{\sqrt{(TP + FP)(TP + FN)(TN + FP)(TN + FN)}}
\end{equation}
Typically, an MCC value between $0.3$-$0.5$ is considered moderate to good performance, and an MCC of $0.5$ or more is considered good to strong performance. Figure~\ref{fig:probe-necessity} shows the MCC of probes trained at each $(t, l)$ position for all four models on Arithmetic and TruthfulQA.

\textbf{Linear separability of necessity is strongly task-dependent.}
Under our model-adaptive definition, the prior ``almost linearly separable'' picture partially holds. On Arithmetic, necessity is linearly separable for most models, with broad regions of mid-to-late layers crossing $\mathrm{MCC} = 0.4$. This aligns with the finding in prior works~\citep{li2025adaptive,wang2026asatrainingfreerepresentationengineering}. On TruthfulQA, however, the regions where MCC exceeds $0.4$ is noticeably smaller, with only near-last tokens in mid-late layers of Llama models still display decent separability. This contrast suggests the challenge in distinguishing model-adaptive \emph{tool-necessary} and \emph{tool-unnecessary} samples, which is more nuanced than the obvious cases prior work focus on~\citep{huang2024metatoolbenchmarklargelanguage,li2025adaptive,wang2026asatrainingfreerepresentationengineering}.
It also suggests that tool-necessity signals are easier to surface in tasks where problem difficulty is reflected in the input’s surface structure, such as arithmetic, where complexity grows with the expression itself. In open-domain factual QA, however, surface form provides little cue about underlying difficulty, making tool necessity or epistemic uncertainty  harder to linearly separate. The heatmap structure also appears similar within model families, with two Qwen and two Llama models sharing similar patterns respsectively.


\textbf{Decent internal signal coexists with large end-to-end mismatch.}
The probe reaches decent MCC at many $(t, l)$ positions, indicating that information about the model's capability boundary is in fact present in the residual stream. Yet the same models still exhibit substantial end-to-end necessity--action mismatch (Table~\ref{tab:call_categories}), meaning this internal signal is not effectively converted into the right tool-call decision at generation time. This mismatch between ``what the hidden states know'' and ``what the model does'' is a first hint of a knowing--doing gap, and motivates the next two questions: does the model encode its action in a similarly separable way (Section~\ref{sec:probe_action}), and how does the action representation relate to the cognition representation (Section~\ref{sec:cognition_execution_gap})?

\subsection{Probing for action}
\label{sec:probe_action}
\begin{figure}[h]
    \centering
    \includegraphics[width=1\linewidth]{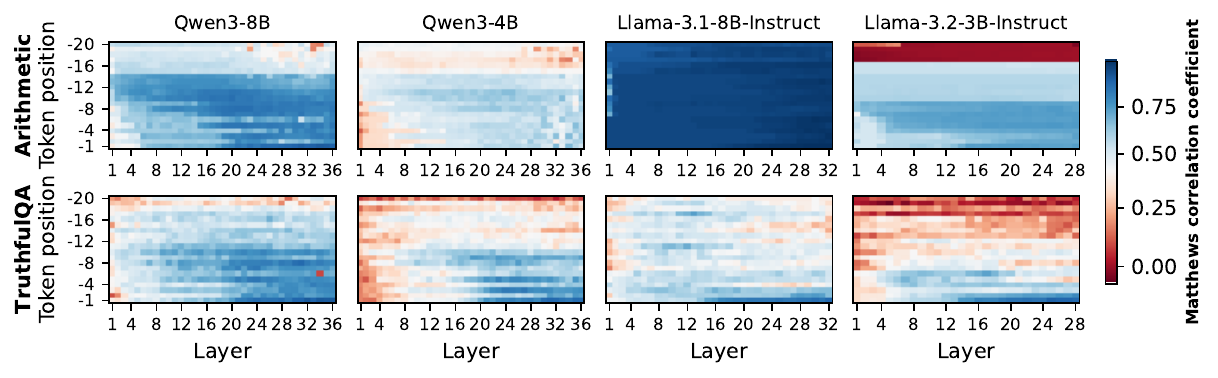}
    \vspace{-0.3cm}
    \caption{\textbf{The action Probe performance on different position in Matthews correlation coefficient.} Each cell reports the held-out MCC of a linear probe trained to predict the tool-call action from the hidden state at a given layer and token position. Darker blue indicates stronger linear separability. The action signal appears highly linearly separable in the hidden states, particularly in near-end tokens and late layers.}
    \label{fig:probe-action}
\end{figure}

Having characterized how necessity is represented internally, we now ask the parallel question for the model's actual decision: how linearly separable is the executed action—whether the model invokes a tool or not—from the same hidden states? Concretely, we train a linear classifier $(\mathbf{w}_a, b_a)$ with the same  objective as in Equation~\ref{eq:probe_objective} while just changing $(\mathbf{w}_c, b_c)$ to $(\mathbf{w}_a, b_a)$.
The Probe performance on the held-out test set (30\% of data) in terms of MCC is shown in Figure~\ref{fig:probe-action}.

\textbf{The action is highly separable from hidden states.}
Figure~\ref{fig:probe-action} shows that, on both Arithmetic and TruthfulQA dataset, the action probe attains $\mathrm{MCC} \geq 0.4$ over broad regions of nearly every model. The signal spans most layers and token positions rather than being confined to a narrow band, indicating that whether the model is about to call a tool is a strongly decodable feature from its residual stream, aligning with recent finding~\citep{esakkiraja2026iamithink}. In Appendix~\ref{append:causal_steering}, we further show that one can causally flip the action via activation steering.

\subsection{The gap between cognition and execution}
\label{sec:cognition_execution_gap}

The two probes give us, at every $(t, l)$, a pair of direction vectors: $\mathbf{w}_c$ pointing from ``unnecessary'' to ``necessary'' in the model's representation space, and $\mathbf{w}_a$ pointing from ``no-call'' to ``call.'' If tool-use behavior were a direct readout of the model's internal necessity assessment, the two directions should align—at least in the layers where both probes succeed. We test this by computing the cosine similarity $\mathrm{CosSim}(\mathbf{w}_c, \mathbf{w}_a)$ between $\mathbf{w}_c$ and $\mathbf{w}_a$ at each position: a value near $\pm 1$ means necessity and action are encoded along (anti-)parallel directions, while a value near $0$ means the two are represented in geometrically independent subspaces.

\begin{figure}[h]
    \centering
    \includegraphics[width=1\linewidth]{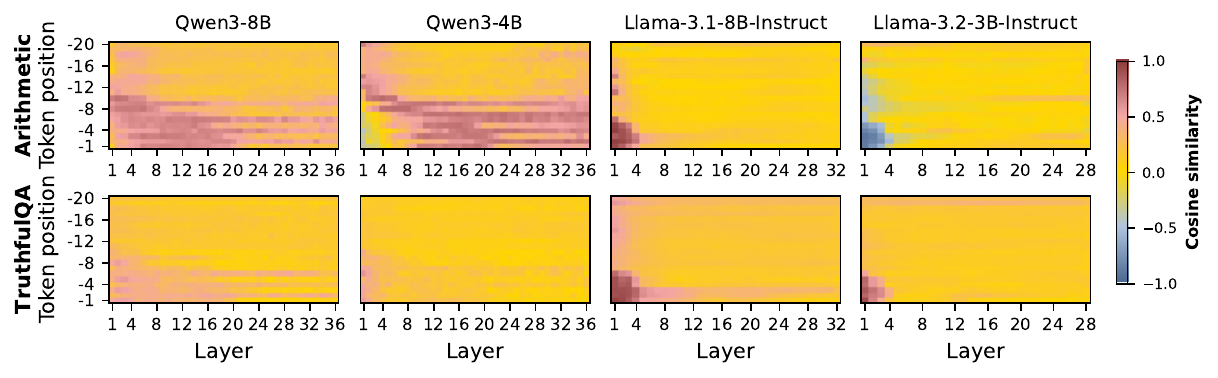}
    \vspace{-0.4cm}
    \caption{\textbf{The cosine similarity score between $
    \mathbf{w_c}$ and $\mathbf{w_a}$ on different positions.} The similarity scores between two probe direction are small across the majority of the area. Although for some models there are moderate similarity scores in late token and middle layer position, two directions fall back to near orthogonal relationship in the late layer of the last token (bottom right corner).}
    \label{fig:probe-cosine}
\end{figure}

\textbf{Partial alignment between $\mathbf{w}_c$ and $\mathbf{w}_a$ exists in intermediate token-layer positions.}
Figure~\ref{fig:probe-cosine} shows that the cosine similarity between $\mathbf{w}_c$ and $\mathbf{w}_a$ is fairly high in notable regions of several heatmaps, particularly covering a sizable area for the two Qwen models on Arithmetic. So necessity and action are not encoded in entirely disjoint subspaces: in some intermediate token-layer positions, the directions share meaningful alignment.

\textbf{Alignment collapses at the position that drives generation.}
The picture changes sharply at the position that actually determines the next token: the late layers of the final query token ($t = -1$, large $l$). For the two models with the strongest mid-stream alignment, Qwen3-8B and Qwen3-4B, the cosine similarity falls back to small values exactly in the bottom right corner of the heatmap, so $\mathbf{w}_c$ and $\mathbf{w}_a$ become close to orthogonal precisely where they would need to interact to translate ``I should call a tool'' into the actual call token. The same trend toward low cosine at late layers / last token holds, more uniformly, for the other models and for TruthfulQA. Whatever partial coupling exists in earlier layers therefore does not survive to the readout. In Appendix~\ref{app:larger-models}, we further show that this decoupling effect also exists in models with much larger scale.

The previous two subsections established that the model's hidden states often contain a usable necessity signal yet still produce mismatched tool-call actions. Figure~\ref{fig:probe-cosine} explains \emph{why}: even when necessity and action share some structure in intermediate representations, the two directions become nearly orthogonal in the late-layer / last-token regime that ultimately drives the next-token decision. 


\subsection{Two stage error diagnosis and attribution}
\label{sec:2_stage_error_diagnosis}

\begin{figure}[h]
    \centering
    \includegraphics[width=1\linewidth]{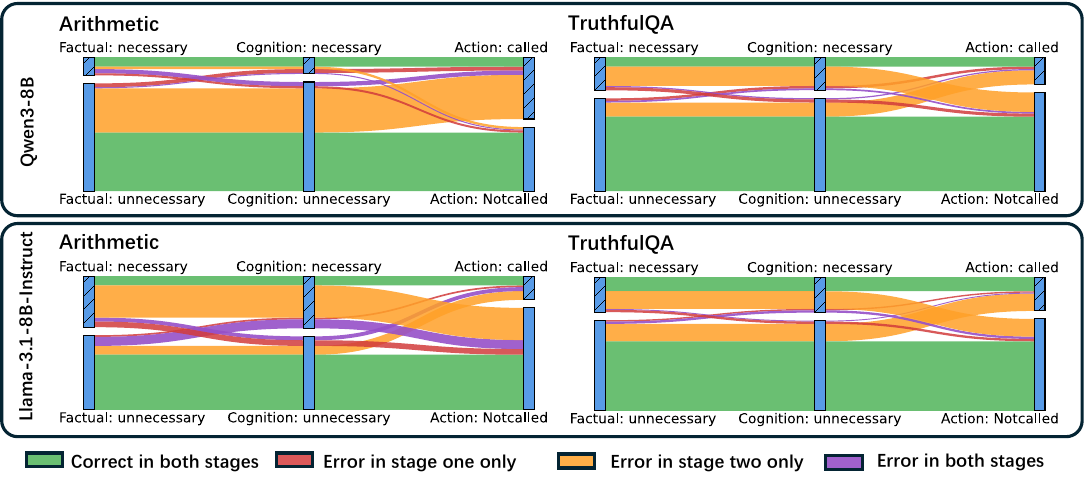}
    \vspace{-0.3cm}
    \caption{\textbf{Per-sample two-stage decomposition of tool-call behavior on Arithmetic and TruthfulQA, for Qwen3-8B (top) and Llama-3.1-8B-Instruct (bottom).} Each flow tracks a sample through three nodes: ground-truth necessity  (\emph{Factual}), the model's internal cognition of necessity (\emph{Cognition}), and the executed action (\emph{Action}). The end-to-end error is dominated by \textcolor{orange!85!black}{\textbf{orange}} flow, where cognition is correct but action flips away from it---the knowing--doing gap.}
    \label{fig:sankey}
\end{figure}

So far we have established two facts: end-to-end necessity--action mismatch is substantial (Table~\ref{tab:call_categories}), and the cognition and action directions are nearly orthogonal at the readout (Section~\ref{sec:cognition_execution_gap}). However, the results in Section~\ref{sec:cognition_execution_gap} tells us only that the two stages are \emph{decoupled}, not which of them is responsible for the mismatch we observe. To attribute the error, we trace each sample along the \textit{Factual~$\rightarrow$~Cognition~$\rightarrow$~Action} modeling, taking Cognition to be the necessity probe $(\mathbf{w}_c, b_c)$ read out at the last query token and last layer---the same position that drives the next-token decision. Each sample then falls into one of four categories: correct in both stages (green), stage-one-only error (red), stage-two-only error (orange, the knowing--doing gap), or compensating errors that cancel at the action (purple). We show the full Sankey flow diagram in Figure~\ref{fig:sankey}.

\textbf{Stage two carries the majority of error.}
In all four panels of Figure~\ref{fig:sankey}, the orange flow (samples with stage two error only) is by far the largest error category, while red (samples with stage one error only) is rather thin. 
Given that cognition and action are decoupled, the asymmetric orange~$\gg$~red localizes the failure: the end-to-end mismatch is overwhelmingly produced in the cognition~$\rightarrow$~action stage rather than in forming cognition itself. The bottleneck is therefore not knowing whether a tool is needed, but converting that knowledge into the call/no-call action. This shows that making correct \textit{when} to use tools decisions is not just about having the correct tool-necessity cognition, but more importantly also about translating that cognition to actual matching action.

\begin{figure}[h]
    \centering
    \includegraphics[width=1\linewidth]{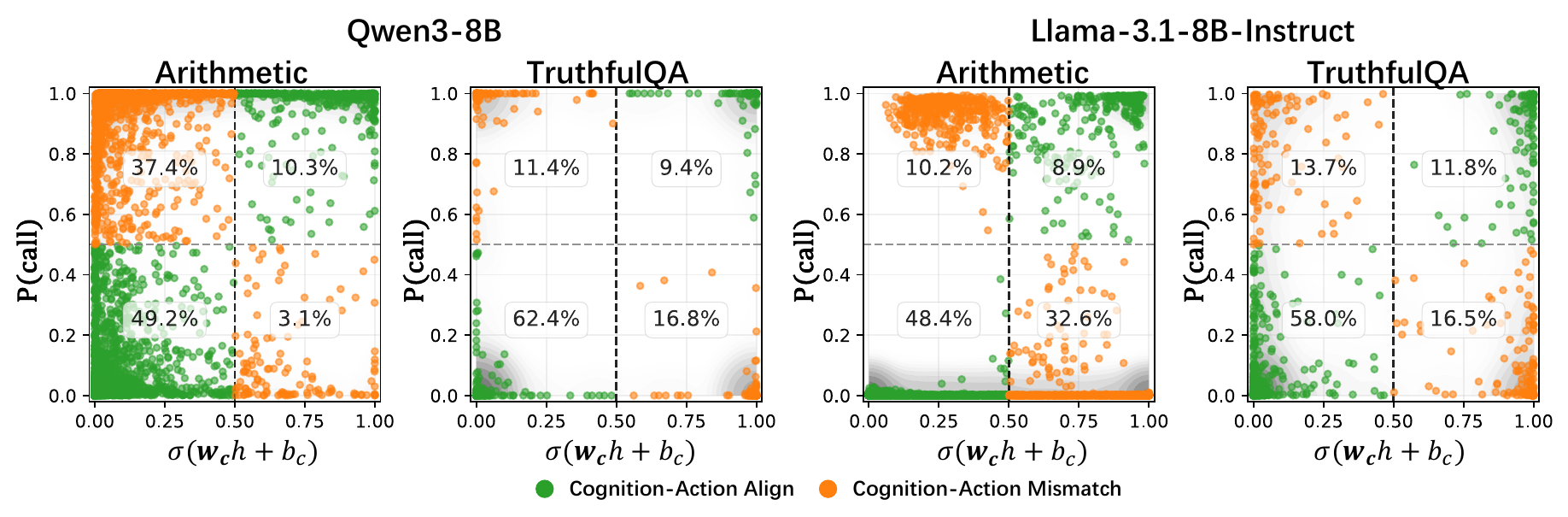}
    \vspace{-0.3cm}
    \caption{\textbf{The confidence of cognition versus tool calling behavior.} The x axis denotes the probe output after sigmoid function, and the y axis represents the tool call probability quantified by Equation~\ref{eq:p_call}. The mismatch can persist even when the internal representation strongly indicates that a sample is either \emph{tool-necessary} or \emph{tool-unnecessary}.}
    \vspace{-2pt}
\label{fig:diagnosis_density_plot}
\end{figure}

\paragraph{The cognition--execution mismatch is not associated with cognition confidence.}
Given the substantial gap between a model's internal cognition and its final tool-call behavior, illustrated by the large orange band in Figure~\ref{fig:sankey}, a natural question is whether this mismatch is caused by uncertainty in the meta-cognitive belief itself. In other words, does the mismatch primarily occur on samples where the model is uncertain about whether a tool is necessary?
To investigate this question, we quantify the confidence of meta-cognition using the post-sigmoid output of the cognition probe, $\sigma(\mathbf{w}_c h + b_c)$. We then plot, for all samples, the relationship between the ``confidence of tool necessity'' and the ``probability of making a tool call'' in Figure~\ref{fig:diagnosis_density_plot}. The probability of making a tool call is defined as
\begin{equation}
\mathrm{P}(\text{call})=
\frac{\mathrm{p}(\langle \text{tool-token} \rangle)}
{\mathrm{p}(\langle \text{tool-token} \rangle)+
\mathrm{p}(\text{best non-tool token})},
\label{eq:p_call}
\end{equation}
where $p(\cdot)$ denotes the softmax probability assigned by the language model to a candidate next token, $\langle \text{tool-token} \rangle$ denotes the model-specific token that initiates a tool call (e.g., \texttt{<tool\_call>} in Qwen models), and the ``best non-tool token'' refers to the highest-logit token among all tokens that are not tool-call tokens. This formulation nicely normalizes the probability to the range $[0,1]$. Since greedy decoding is used when collecting tool-call behaviors (Section~\ref{sec:collect_tool_Call_behav}), a value of $P(\text{call}) > 0.5$ corresponds to an actual tool call being generated.
As shown in Figure~\ref{fig:diagnosis_density_plot}, the cognition--execution mismatch does not primarily occur near the uncertain region where $\sigma(\mathbf{w}_c h + b_c) \approx 0.5$. 
Instead, many orange points occur in regions where $\sigma(\mathbf{w}_c h + b_c)$ is close to $0$ or $1$. 
This observation suggests that the cognition--execution mismatch is not driven by low confidence in the model's internal cognition. Rather, the mismatch can persist even when the internal representation strongly indicates that a sample is either \emph{tool-necessary} or \emph{tool-unnecessary}.

\paragraph{What can we do about it?}
A natural question arising from our diagnosis is how to address the identified knowing-doing gap. The most straightforward mitigation solution that naturally extends from our results is to use the trained necessity probe as a latent tool-necessity estimator during inference and adjust the model's tool-calling behavior on the fly, either through activation steering or guided decoding. This directly targets both types of misaligned tool-use cases: \emph{Necessary-NotCalled} and \emph{Unnecessary-Called}.

Since the probe is trained on the training split, evaluating the mitigation on that same split would not be meaningful. We therefore conduct additional mitigation experiments on Qwen3-8B and Llama-3.1-8B-Instruct using held-out test sets. When the probe predicts ``necessary,'' we apply guided decoding to force the tool-call token and trigger a tool call; when the probe predicts ``unnecessary,'' we instead mask the tool-call token to force a direct answer. The results are shown in Table~\ref{tab:probe_mitigation}.

\begin{table*}[ht]
\centering
\small
\setlength{\tabcolsep}{8pt}
\caption{Mitigation of misaligned tool use through probe-guided decoding on held-out test sets. Misaligned rates are reported before and after mitigation, along with relative reductions.}
\label{tab:probe_mitigation}
\vspace{6pt}
\resizebox{\textwidth}{!}{%
\begin{tabular}{llccc}
\toprule
\textbf{Model}
& \textbf{Dataset}
& \shortstack{\textbf{Misaligned Rate} \\ \textbf{Before}}
& \shortstack{\textbf{Misaligned Rate} \\ \textbf{After}}
& \shortstack{\textbf{Relative} \\ \textbf{Reduction}} \\
\midrule
\multirow{2}{*}{Qwen3-8B}
& Arithmetic  & 41.58\% & 9.42\%  & 77.35\% \\
& TruthfulQA  & 30.49\% & 19.92\% & 34.67\% \\
\midrule
\multirow{2}{*}{Llama-3.1-8B-Instruct}
& Arithmetic  & 37.83\% & 14.75\% & 61.01\% \\
& TruthfulQA  & 28.66\% & 15.85\% & 44.68\% \\
\bottomrule
\end{tabular}%
}
\end{table*}

Across both models and datasets, leveraging the necessity probe as an inference-time control signal consistently reduces misaligned tool use, with relative reductions ranging from 34.67\% to 77.35\%. These results provide a simple and concrete mitigation strategy that naturally follows from our analysis, demonstrating that the latent necessity signal identified in our work can be operationalized to improve tool-calling behavior on unseen examples.

\section{Conclusion}
In this work, we introduced a model-adaptive definition of tool necessity that grounds evaluation in empirical capabilities, and revealed a substantial mismatch between when models actually need tools and when they invoke them. By decomposing the tool-use process into internal cognition and execution stages, and analyzing hidden state representations, we identified a fundamental "knowing-doing gap" in LLMs. While models sometimes internally recognize the tool necessity, these cognitive representations become orthogonally misaligned with execution intent in later layers, leading to failures in taking the appropriate action. Our findings demonstrate that improving autonomous agents requires not just better internal meta-cognition, but bridging the knowing-doing gap to ensure self-awareness translates into reliable execution.

\section*{Acknowledgments}

This work was supported in part by NSF CAREER Award 1942230, the ONR PECASE Award N00014-25-1-2378, ARO Early Career Program Award 310902-00001, Army Grant W911NF-21-2-0076, NSF Award CCF-2212458, NSF Award 2229885 (NSF Institute for Trustworthy AI in Law and Society, TRAILS), MURI Grant 14262683, DARPA AIQ Grant HR00112590066, and a Meta Research Award 314593-00001.

\bibliographystyle{plainnat}
\bibliography{custom}

\newpage
\appendix

\section{More details on arithmetic dataset curation}
\label{app:arithmetic}
We generate math arithmetic problems grouped into three types. The easy group contains one-step addition/subtraction, short two-step chains, and small modulo problems. These give cases where a calculator is usually not needed. The larger short group keeps the expressions simple, but uses larger operands, including multi-digit subtraction, four-digit addition/subtraction, and two- or three-digit multiplication. These problems are still short enough to invite a direct answer, but they are more likely to cause digit errors. The multi-step group contains precedence chains, parenthesized expressions, multiplication chains, and long addition/subtraction chains. These examples test whether the model can track intermediate values and apply the order of operations, especially in cases that look simple but are easy to miscompute. We sample the dataset with a fixed random seed. During generation, we skip repeated expressions and resample until each family reaches its assigned sampling share. Table~\ref{table:math-arithmetic} gives the problem families and their sampling shares.

\begin{table}[htbp]
\centering
\caption{Breakdown of the math arithmetic dataset. Shares show the fixed sampling share for each problem family. We use $4000$ samples overall.}
\label{table:math-arithmetic}
\begin{adjustbox}{width=\linewidth}

\begin{tabular}{l l r p{0.40\textwidth} l}
\toprule
\textbf{Group} & \textbf{Problem Family} & \textbf{Share} & \textbf{Detail} & \textbf{Example} \\
\midrule
\multirow{3}{*}{Easy}
& Single-step arithmetic & 8\%
& Two operands from 1--99 with an addition or subtraction operator.
& \texttt{21 + 59} \\

& Two-step arithmetic & 5\%
& Three operands from 1--99 with addition or subtraction operators.
& \texttt{70 - 47 + 68} \\

& Small modulo & 5\%
& Three-digit dividend with divisor from 3--19.
& \texttt{562 \% 8} \\

\midrule

\multirow{5}{*}{Larger short}
& Negative subtraction & 7\%
& Three- or four-digit operand minus a larger operand.
& \texttt{390 - 554} \\

& Four-digit addition/subtraction & 6\%
& Two four-digit operands.
& \texttt{4921 - 9108} \\

& Two-digit multiplication & 9\%
& Two two-digit operands.
& \texttt{34 * 75} \\

& Three-by-two multiplication & 9\%
& One three-digit factor and one two-digit.
& \texttt{504 * 61} \\

& Three-by-three multiplication & 7\%
& Two three-digit factors.
& \texttt{867 * 671} \\

\midrule

\multirow{5}{*}{Multi-step}
& Precedence chain & 12\%
& Five two- or three-digit operands with addition, subtraction, or multiplication operators.
& \texttt{84 * 82 - 755 - 805 - 29} \\

& One-digit addition/subtraction chain & 11\%
& 16--39 one-digit terms with addition or subtraction operators.
& \texttt{3 + 4 + 1 + 3 - 8 - \dots} \\

& Small addition/subtraction chain & 10\%
& 21--27 terms from 1--30 with addition or subtraction operators.
& \texttt{9 + 5 - 1 - 3 + 23 + \dots} \\

& Parenthesized expression & 6\%
& Four two-digit operands in $(a+b)\times(c-d)$ form.
& \texttt{(67 + 68) * (52 - 88)} \\

& Multiplication chain & 5\%
& Five two-digit operands in $a+b\times c-d\times e$ form.
& \texttt{94 + 40 * 50 - 24 * 87} \\
\bottomrule
\end{tabular}
\end{adjustbox}
\end{table}

Algorithms~\ref{alg:single-step-arithmetic}--\ref{alg:multiplication-chain} specify the exact procedures used for generating the data samples in each family.
In these algorithms, $\mathcal{U}\{m,\ldots,n\}$ denotes the discrete uniform distribution over integers from $m$ to $n$, and $\mathcal{U}(0,1)$ denotes the continuous uniform distribution on the unit interval.

\begin{algorithm}[H]
\caption{\textsc{SingleStepArithmetic}}
\label{alg:single-step-arithmetic}
\begin{algorithmic}[1]
\State $a,b \sim \mathcal{U}\{1,\ldots,99\}$
\State $op \sim \{+,-\}$
\State \Return ``$a\ op\ b$''
\end{algorithmic}
\end{algorithm}

\begin{algorithm}[H]
\caption{\textsc{TwoStepArithmetic}}
\label{alg:two-step-arithmetic}
\begin{algorithmic}[1]
\State $a,b,c \sim \mathcal{U}\{1,\ldots,99\}$
\State $u \sim \mathcal{U}(0,1)$
\If{$u < 0.5$}
    \State \Return ``$a + b - c$''
\Else
    \State \Return ``$a - b + c$''
\EndIf
\end{algorithmic}
\end{algorithm}

\begin{algorithm}[H]
\caption{\textsc{SmallModulo}}
\label{alg:small-modulo}
\begin{algorithmic}[1]
\State $a \sim \mathcal{U}\{100,\ldots,999\}$
\State $b \sim \mathcal{U}\{3,\ldots,19\}$
\State \Return ``$a\ \%\ b$''
\end{algorithmic}
\end{algorithm}

\begin{algorithm}[H]
\caption{\textsc{NegativeSubtraction}}
\label{alg:negative-subtraction}
\begin{algorithmic}[1]
\State $u \sim \mathcal{U}(0,1)$
\If{$u < 0.55$}
    \State $a \sim \mathcal{U}\{100,\ldots,500\}$
    \State $b \sim \mathcal{U}\{a+10,\ldots,a+250\}$
\Else
    \State $a \sim \mathcal{U}\{1000,\ldots,5000\}$
    \State $b \sim \mathcal{U}\{a+100,\ldots,a+3000\}$
\EndIf
\State \Return ``$a - b$''
\end{algorithmic}
\end{algorithm}

\begin{algorithm}[H]
\caption{\textsc{FourDigitAdditionSubtraction}}
\label{alg:four-digit-add-sub}
\begin{algorithmic}[1]
\State $a,b \sim \mathcal{U}\{1000,\ldots,9999\}$
\State $u \sim \mathcal{U}(0,1)$
\If{$u < 0.6$}
    \State \Return ``$a + b$''
\Else
    \State \Return ``$a - b$''
\EndIf
\end{algorithmic}
\end{algorithm}

\begin{algorithm}[H]
\caption{\textsc{TwoDigitMultiplication}}
\label{alg:two-digit-multiplication}
\begin{algorithmic}[1]
\State $u \sim \mathcal{U}(0,1)$
\If{$u < 0.45$}
    \State $a,b \sim \mathcal{U}\{15,\ldots,50\}$
\Else
    \State $a,b \sim \mathcal{U}\{30,\ldots,99\}$
\EndIf
\State \Return ``$a \times b$''
\end{algorithmic}
\end{algorithm}

\begin{algorithm}[H]
\caption{\textsc{ThreeByTwoMultiplication}}
\label{alg:three-by-two-multiplication}
\begin{algorithmic}[1]
\State $a \sim \mathcal{U}\{100,\ldots,999\}$
\State $b \sim \mathcal{U}\{10,\ldots,99\}$
\State \Return ``$a \times b$''
\end{algorithmic}
\end{algorithm}

\begin{algorithm}[H]
\caption{\textsc{ThreeByThreeMultiplication}}
\label{alg:three-by-three-multiplication}
\begin{algorithmic}[1]
\State $a,b \sim \mathcal{U}\{100,\ldots,999\}$
\State \Return ``$a \times b$''
\end{algorithmic}
\end{algorithm}

\begin{algorithm}[H]
\caption{\textsc{PrecedenceChain}}
\label{alg:precedence-chain}
\begin{algorithmic}[1]
\State $a_1,\ldots,a_5 \sim \mathcal{U}\{10,\ldots,999\}$
\State $op_1,\ldots,op_4 \sim \{+,-,\times\}$
\State \Return ``$a_1\ op_1\ a_2\ op_2\ a_3\ op_3\ a_4\ op_4\ a_5$''
\end{algorithmic}
\end{algorithm}

\begin{algorithm}[H]
\caption{\textsc{OneDigitAdditionSubtractionChain}}
\label{alg:one-digit-chain}
\begin{algorithmic}[1]
\State $u \sim \mathcal{U}(0,1)$
\If{$u < 0.4$}
    \State $n \sim \mathcal{U}\{16,\ldots,22\}$
\Else
    \State $n \sim \mathcal{U}\{29,\ldots,39\}$
\EndIf
\State $a_1,\ldots,a_n \sim \mathcal{U}\{1,\ldots,9\}$
\State For each $i$, sample $op_i$ with $\Pr(op_i=+)=0.53$ and $\Pr(op_i=-)=0.47$
\State \Return ``$a_1\ op_1\ a_2\ op_2\ \cdots\ op_{n-1}\ a_n$''
\end{algorithmic}
\end{algorithm}

\begin{algorithm}[H]
\caption{\textsc{SmallAdditionSubtractionChain}}
\label{alg:small-chain}
\begin{algorithmic}[1]
\State $n \sim \mathcal{U}\{21,\ldots,27\}$
\State $a_1,\ldots,a_n \sim \mathcal{U}\{1,\ldots,30\}$
\State $op_1,\ldots,op_{n-1} \sim \{+,-\}$
\State \Return ``$a_1\ op_1\ a_2\ op_2\ \cdots\ op_{n-1}\ a_n$''
\end{algorithmic}
\end{algorithm}

\begin{algorithm}[H]
\caption{\textsc{ParenthesizedExpression}}
\label{alg:parenthesized-expression}
\begin{algorithmic}[1]
\State $a,b,c,d \sim \mathcal{U}\{10,\ldots,99\}$
\State \Return ``$(a+b)\times(c-d)$''
\end{algorithmic}
\end{algorithm}

\begin{algorithm}[H]
\caption{\textsc{MultiplicationChain}}
\label{alg:multiplication-chain}
\begin{algorithmic}[1]
\State $a,b,c,d,e \sim \mathcal{U}\{10,\ldots,99\}$
\State \Return ``$a + b \times c - d \times e$''
\end{algorithmic}
\end{algorithm}

\section{Sensitivity analysis under alternative $N$, $T$ values}
\label{append:alternative_NT}

We conducted sensitivity analysis using
$N \in \{5, 10, 15\}$ at $T = 0.7$ and
$T \in \{0.5, 0.7, 0.9\}$ at $N = 10$,
across both datasets and all four models.
As shown in Table~\ref{tab:sensitivity_NT},
increasing $N$ consistently increased the proportion
of tool-necessary samples in every dataset-model
combination: averaged across models, the proportion
increased from 22.5\% to 28.2\% to 31.4\% on
TruthfulQA, and from 26.2\% to 32.0\% to 34.7\%
on arithmetic.
Increasing $T$ produced the same overall tendency,
although the effect was smaller: the average
tool-necessary proportion changed from 28.0\% to
28.2\% to 28.4\% on TruthfulQA and from 29.9\%
to 32.0\% to 33.3\% on arithmetic.
These results confirm the intended behavior:
stricter settings generally classify more queries
as tool-necessary, while the paper's qualitative
conclusions are not tied to selecting specific
$N$/$T$ configurations.

\begin{table*}[h]
\centering
\small
\setlength{\tabcolsep}{7pt}
\caption{Sensitivity analysis under alternative $N$
and $T$ values across TruthfulQA and arithmetic.
Entries report the percentage of prompts labeled
tool-necessary. The Average column reports the
mean across the four models.}
\vspace{6pt}
\label{tab:sensitivity_NT}

\begin{tabular}{llccccc}
\toprule
\multicolumn{7}{l}{
\textbf{(a) Fixed $T = 0.7$: varying $N$}
} \\
\midrule
\textbf{Dataset} & \textbf{$N$}
& \textbf{Average}
& \textbf{Llama-3.1-8B}
& \textbf{Llama-3.2-3B}
& \textbf{Qwen3-4B}
& \textbf{Qwen3-8B} \\
\midrule
\multirow{3}{*}{TruthfulQA}
& 5  & \textbf{22.5\%} & 20.9\% & 20.0\% & 26.6\% & 22.6\% \\
& 10 & \textbf{28.2\%} & 27.9\% & 27.2\% & 31.3\% & 26.3\% \\
& 15 & \textbf{31.4\%} & 32.6\% & 33.3\% & 32.4\% & 27.1\% \\
\midrule
\multirow{3}{*}{Arithmetic}
& 5  & \textbf{26.2\%} & 33.9\% & 46.7\% & 13.9\% & 10.2\% \\
& 10 & \textbf{32.0\%} & 40.6\% & 52.1\% & 20.9\% & 14.4\% \\
& 15 & \textbf{34.7\%} & 43.8\% & 53.9\% & 24.6\% & 16.3\% \\
\midrule
\multicolumn{7}{l}{
\textbf{(b) Fixed $N = 10$: varying $T$}
} \\
\midrule
\textbf{Dataset} & \textbf{$T$}
& \textbf{Average}
& \textbf{Llama-3.1-8B}
& \textbf{Llama-3.2-3B}
& \textbf{Qwen3-4B}
& \textbf{Qwen3-8B} \\
\midrule
\multirow{3}{*}{TruthfulQA}
& 0.5 & \textbf{28.0\%} & 27.4\% & 28.5\% & 29.6\% & 26.4\% \\
& 0.7 & \textbf{28.2\%} & 27.9\% & 27.2\% & 31.3\% & 26.3\% \\
& 0.9 & \textbf{28.4\%} & 27.4\% & 27.9\% & 31.3\% & 27.1\% \\
\midrule
\multirow{3}{*}{Arithmetic}
& 0.5 & \textbf{29.9\%} & 37.4\% & 49.6\% & 19.6\% & 13.1\% \\
& 0.7 & \textbf{32.0\%} & 40.6\% & 52.1\% & 20.9\% & 14.4\% \\
& 0.9 & \textbf{33.3\%} & 43.5\% & 53.7\% & 21.8\% & 14.2\% \\
\bottomrule
\end{tabular}
\end{table*}

\section{Causally flipping actions via activation steering on the action probe}
\label{append:causal_steering}

To investigate causality, we conduct activation steering to examine whether intervening on the model's hidden activations can flip its tool-use actions. Specifically, we intervene on the model's hidden activations at inference time and test whether moving the representation along the action probe direction changes the resulting tool-use action.

We evaluate Llama-3.1-8B-Instruct and Qwen3-8B on both Arithmetic and TruthfulQA, applying steering to the originally misaligned cases: positive steering is applied to Necessary-NotCalled examples to induce a tool call, while negative steering is applied to Unnecessary-Called examples to suppress an unnecessary call. Thus, a successful flip corresponds to correcting the model's original behavior.

After sweeping over a few possible steering strengths, we obtain the corrective flip rates shown in Table~\ref{tab:causal_steering}. The results demonstrate that steering along the action probe direction can effectively alter the model's tool-use decisions, providing causal evidence that the probed action representation is involved in the model's tool-use behavior. Such steering would not be meaningful on the cognition probe as only the action is directly observable from the model output. There exists no observable model output that can faithfully represent a model's internal cognition, hence the probing result on the cognition probe remains correlational.

\begin{table*}[ht]
\centering
\small
\setlength{\tabcolsep}{9pt}
\caption{Corrective action flip rates (\%) under activation steering along the action probe direction. Negative steering is applied to Unnecessary-Called examples to suppress unnecessary tool calls, while positive steering is applied to Necessary-NotCalled examples to induce necessary tool calls.}
\vspace{6pt}
\label{tab:causal_steering}

\resizebox{\textwidth}{!}{%
\begin{tabular}{llcc}
\toprule
\textbf{Dataset}
& \textbf{Model}
& \shortstack{\textbf{Suppress unnecessary calls} \\ \textbf{(negative direction)}}
& \shortstack{\textbf{Induce necessary calls} \\ \textbf{(positive direction)}} \\
\midrule
\multirow{2}{*}{Arithmetic}
& Llama-3.1-8B-Instruct & 100.0\% & 100.0\% \\
& Qwen3-8B & 94.5\% & 100.0\% \\
\midrule
\multirow{2}{*}{TruthfulQA}
& Llama-3.1-8B-Instruct & 99.1\% & 98.4\% \\
& Qwen3-8B & 97.0\% & 66.2\% \\
\bottomrule
\end{tabular}%
}
\end{table*}

\section{Generalization to larger models}
\label{app:larger-models}

To test whether the knowing--doing gap persists at a substantially larger model scale, we repeat our probing analysis on Llama-3.1-70B-Instruct~\cite{grattafiori2024llama}. We use the same definition of model-adaptive necessity labels as described in Section~\ref{sec:necessary_and_unnecessary} and linear-probe protocol as in Sections~\ref{sec:probe_cognition} and \ref{sec:probe_action}. For each dataset, we train a cognition probe to predict whether tool use is necessary for the model and an action probe to predict whether the model calls the tool. We report held-out Matthews correlation coefficient (MCC) and compute the cosine similarity between the two probes' weight vectors at each token--layer position. The 70B activations cover all 80 transformer layers and the final five query tokens.

\begin{figure*}[ht]
    \centering
    \includegraphics[width=\textwidth]{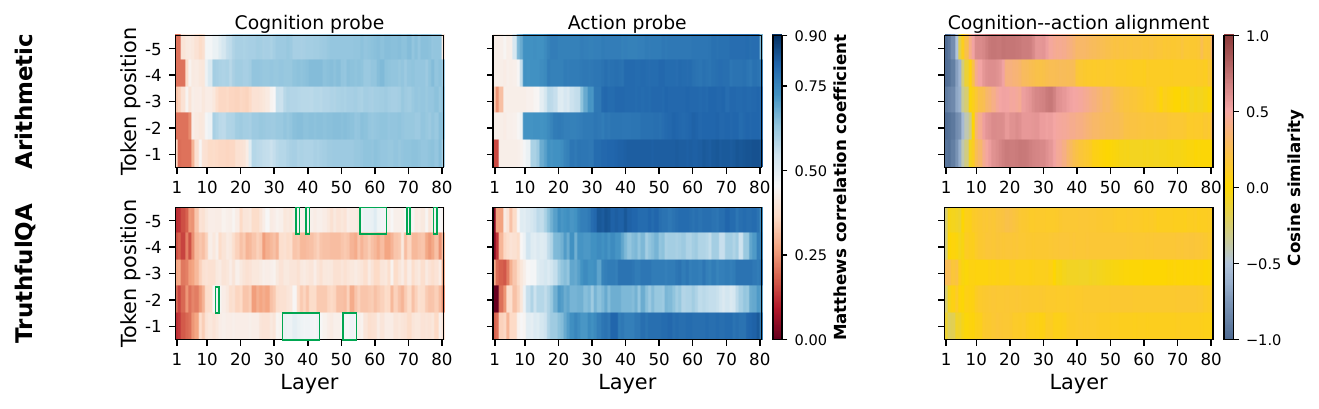}
    \vspace{-8pt}
    \caption{\textbf{Cognition, action, and cognition--action alignment for Llama-3.1-70B-Instruct.} Rows show Arithmetic (top) and TruthfulQA (bottom); columns show the held-out MCC of the cognition (necessity) probe, the held-out MCC of the action probe, and the cosine similarity between their weight vectors. Each heatmap covers all 80 layers and the final five query tokens. Green outlines in the TruthfulQA cognition panel mark contiguous regions in which MCC $\geq 0.45$.}
    \label{fig:llama31-70b-probes}
\end{figure*}

\paragraph{Necessity remains decodable at 70B, with the same task dependence.}
Figure~\ref{fig:llama31-70b-probes} shows a broad and stable necessity signal on Arithmetic: 86.25\% of the evaluated token--layer positions attain cognition-probe MCC $\geq 0.4$. The signal is weaker and more localized on TruthfulQA, where 39.50\% of positions cross the same threshold. Nevertheless, the meta-cognitive tool-necessity signal remains present at many token--layer positions: the green boxes in the TruthfulQA cognition heatmap enclose 28 positions at which MCC $\geq 0.45$. Necessity is therefore still meaningfully decodable in the 70B model on TruthfulQA, although less broadly and less strongly than the model's action. Moreover, the lower cognition-probe performance on TruthfulQA than on Arithmetic is consistent with the results for the other models in Section~\ref{sec:probe_cognition}. Increasing model scale thus does not remove the domain dependence observed in the main experiments: model-adaptive necessity is easier to recover from arithmetic representations than from open-domain factual questions.

\paragraph{Action is consistently separable across layer--token positions.}
The action probe achieves MCC $\geq 0.4$ at 98.75\% of positions on Arithmetic and 90.25\% on TruthfulQA. The signal spans most layers and token positions rather than being confined to a localized region, indicating that whether the model is about to call a tool is a strongly decodable feature from its residual stream, aligning with the findings in Section~\ref{sec:probe_action}.

\paragraph{Cognition and action become nearly orthogonal at the readout.}
Although the Arithmetic heatmap contains positive alignment in earlier layers, this alignment does not persist to the position that directly drives generation. At the final token and final layer, the cognition--action cosine similarity is $-0.018$ on Arithmetic and $0.026$ on TruthfulQA. Averaged over the final token in layers 61--80, the corresponding similarities are $-0.041$ and $0.078$. Hence, the larger model reproduces the geometry found in our main results: both signals can be linearly recoverable while their directions are nearly orthogonal at the output-facing representation. These results suggest that the knowing--doing gap is not confined to the smaller models studied in our main experiments.

\section{Explicitly prompting for verbalized belief of tool-necessity}
\label{append:verbal_tool_necessity}
Due to the limitation of LLMs in verbalizing internal decision processes~\citep{lindsey2025biology,jian2025languagemodelscapablemetacognitive}, and the fundamental difference between the task of self-assessment and actual problem solving, in this paper, we followed the approach in recent work that use internal state probing to measure models' cognition of tool-necessity~\citep{li2025adaptive,wang2026asatrainingfreerepresentationengineering}. Nevertheless, for completeness, we also report results obtained using explicit self-assessment prompts.

Specifically, we adopt a two-stage inference procedure. In the first stage, the model is given the same questions from the Arithmetic and TruthfulQA datasets, but instead of solving the problem directly, it is prompted to first decide ``whether it is necessary to invoke an external tool'' and to answer only with `yes' or `no'. In the second stage, the model is instructed to ``Now answer the original user request.''

Table~\ref{tab:mcc_mismatch_changed} reports: (1) the MCC between the model's `yes'/`no' responses and the actual capability-grounded tool necessity defined in Section~\ref{sec:necessary_and_unnecessary}; (2) the cognition--execution mismatch rate, defined as the proportion of samples where the model answered `yes' but did not invoke a tool, or answered `no' but eventually invoked one; and (3) the proportion of samples whose eventual tool-calling behavior changed relative to the direct task-oriented prompting setup used in Section~\ref{sec:collect_tool_Call_behav}.

The results show that the MCC of explicit `yes'/`no' judgments is substantially worse at capturing the actual capability-grounded notion of tool necessity. In particular, Llama-3.2-3B-Instruct achieves a negative MCC on TruthfulQA, while Llama-3.1-8B-Instruct simply answers `no' for every TruthfulQA sample, resulting in an undefined MCC. This behavior implies that the model judges no sample to require a tool, which is clearly inconsistent with the capability measurements reported in Section~\ref{sec:necessary_and_unnecessary}. This poor MCC results further show the challenge in distinguishing model-adaptive tool-necessary and tool-unnecessary samples, which is more nuanced than the obvious cases prior work focus on~\citep{huang2024metatoolbenchmarklargelanguage,li2025adaptive,wang2026asatrainingfreerepresentationengineering}.

In contrast, the cognition--execution mismatch rates are noticeably lower than those reported in Section~\ref{sec:2_stage_error_diagnosis}. Llama-3.1-8B-Instruct even achieves a 0 mismatch rate, meaning that it not only answered `no' for all samples, but also consistently refrained from making any tool calls. This outcome is somewhat expected: once the `yes'/`no' response becomes part of the model's context, the model is more likely to remain consistent with that earlier commitment during subsequent generation.

Most importantly, however, Table~\ref{tab:mcc_mismatch_changed} shows a large ``Changed'' rate in the third column of each dataset. Relative to direct problem solving with the task-oriented prompts used in our main experiments, explicit self-assessment changes tool-calling behavior on up to nearly 50\% of samples. In practical deployments, prompts are typically task-oriented and designed to maximize task performance, rather than to elicit explicit self-assessment. Therefore, this substantial shift in behavior suggests that evaluations based on explicit prompts such as ``decide whether it is necessary to invoke an external tool and answer `yes' or `no''', as used in some prior work~\citep{huang2024metatoolbenchmarklargelanguage,li2025adaptive}, may produce results that diverge significantly from models' actual tool-use behavior under realistic task settings.

\begin{table}[ht]
\centering
\caption{Tool-call evaluation summary across datasets. For each model and dataset, we report Matthews Correlation Coefficient (MCC), mismatch rate $(\text{yes,false}) + (\text{no,true})$, and the proportion of changed tool-call behavior across variants.}
\label{tab:mcc_mismatch_changed}
\resizebox{\linewidth}{!}{%
\begin{tabular}{lccc ccc}
\toprule
\multirow{2}{*}{Model} & \multicolumn{3}{c}{Arithmetic} & \multicolumn{3}{c}{TruthfulQA} \\
\cmidrule(lr){2-4} \cmidrule(lr){5-7}
 & MCC & Cog-Exe-Mis. (\%) & Changed (\%) & MCC & Cog-Exe-Mis. (\%) & Changed (\%) \\
\midrule
Llama-3.1-8B-Instruct & 0.0712 & 5.42 & 18.20 & n/a & 0.00 & 26.93 \\
Llama-3.2-3B-Instruct & 0.2251 & 19.62 & 29.18 & -0.0400 & 9.67 & 27.05 \\
Qwen3-4B              & 0.2330 & 3.85 & 29.55 & 0.1915 & 36.35 & 34.64 \\
Qwen3-8B              & 0.2016 & 2.12 & 49.27 & 0.0538 & 13.34 & 20.93 \\
\bottomrule
\end{tabular}%
}
\end{table}



\section{Limitations}
Our experimental setting is relatively restricted to Arithmetic and TruthfulQA with simple calculator and search tools, and extending the analysis to long-horizon, multi-turn agentic environments would be valuable. However, the primary goal of this paper is diagnostic and interpretability-oriented: we aim to isolate the relationship among a model’s capability boundary, its internal representation of tool necessity, and its eventual tool-call action. Controlled settings like the ones used allow us to define these quantities cleanly and avoid several confounds that arise in long-horizon tasks.

\end{document}